\documentclass[runningheads]{llncs}
\usepackage[T1]{fontenc}
\usepackage{graphicx}
\usepackage{float}

\usepackage{subfigure}

\usepackage{todonotes}

\begin{document}
\title{The Impact of Prosodic Segmentation on Speech Synthesis of Spontaneous Speech}

\titlerunning{The Impact of Prosodic Segmentation ...}

\author{\textbf{Julio Galdino}\inst{1}\orcidID{0000-0001-6378-4648} \and
\textbf{Sidney Leal}\inst{1,5}\orcidID{0000-0002-8817-2063} \and
Leticia de Souza\inst{2}\orcidID{0009-0009-7191-9296} \and
Rodrigo Lima\inst{1}\orcidID{0009-0009-4344-1109} \and
Antonio Moreira\inst{1}\orcidID{0009-0001-9867-3101} \and
Arnaldo Candido Jr.\inst{2}\orcidID{0000-0002-5647-0891} \and
Miguel Oliveira Jr.\inst{3}\orcidID{0000-0002-0866-0535} \and
Edresson Casanova\inst{4}\orcidID{0000-0003-0160-7173} \and
Sandra Aluísio\inst{1}\orcidID{0000-0001-5108-2630}
}
\institute{University of São Paulo, São Carlos, SP, Brazil 
\and
Universidade Estadual Paulista, São José do Rio Preto, SP, Brazil 
\and
Universidade Federal de Alagoas,  Maceió, AL, Brazil 
\and
NVIDIA Corporation, São Paulo, SP, Brazil 
\and
Venturus - Centro de Inovação Tecnológica, Campinas, SP, Brazil 
\\
\textbf{Corresponding authors:} juliogaldino@usp.br, sidleal@gmail.com
}
\authorrunning{Julio Cesar Galdino et al.}

\maketitle              

\begin{abstract}

Spontaneous speech presents several challenges for speech synthesis, particularly in capturing the natural flow of conversation, including turn-taking, pauses, and disfluencies. Although speech synthesis systems have made significant progress in generating natural and intelligible speech, primarily through architectures that implicitly model prosodic features such as pitch, intensity, and duration, the construction of datasets with explicit prosodic segmentation and their impact on spontaneous speech synthesis remains largely unexplored. This paper evaluates the effects of manual and automatic prosodic segmentation annotations in Brazilian Portuguese on the quality of speech synthesized by a non-autoregressive model, FastSpeech 2. Experimental results show that training with prosodic segmentation produced slightly more intelligible and acoustically natural speech. While automatic segmentation tends to create more regular segments, manual prosodic segmentation introduces greater variability, which contributes to more natural prosody. Analysis of neutral declarative utterances showed that both training approaches reproduced the expected nuclear accent pattern, but the prosodic model aligned more closely with natural pre-nuclear contours. To support reproducibility and future research, all datasets, source codes, and trained models are publicly available under the CC BY-NC-ND 4.0 license. 
\end{abstract}
\section{Introduction}
\label{sec:1}

Text-to-Speech (TTS) systems have achieved significant progress in recent years, particularly in generating natural and intelligible speech from text. Advances in system architectures have enabled fine-grained control over synthesized speech by implicitly modeling various attributes, including emotion and prosodic features such as pitch, intensity, and duration \cite{xie2025controllablespeechsynthesisera}. Additionally, incorporating regional linguistic variations has become increasingly important to enhance the naturalness and authenticity of synthesized voices \cite{matos-etal-2024-accent}.  Nevertheless, there remains a pressing need for conversational and spontaneous speech datasets, which are crucial for training applications such as chatbots, virtual assistants, and interactive storytelling systems \cite{li24na_interspeech}. 

The field of speech synthesis has witnessed significant advances, and research in automatic speech recognition (ASR) has also made notable progress, particularly with the widespread adoption of systems such as OpenAI's Whisper \cite{radford23_whisper} and related developments such as WhisperX \cite{bain2022whisperx}, which enable fast ASR (up to 70× real-time transcription) based on the Whisper large-v2 model. As a result, several datasets for low-resource languages have become available through the use of ASR tools, often supplemented with further human revision \cite{evaldo-leal-etal-2025-mupe}. However, an important concern remains: Are we losing critical prosodic information when using ASR systems to transcribe and segment spontaneous speech?

The relationship between speech synthesis and prosody has been studied for decades. For example, \cite{sagisaka1997computing} presents a collection of papers on computational approaches to processing spontaneous speech prosody, exploring the generation and modeling of prosody in computational synthesis. The evaluation of prosodic aspects of synthesized speech also has a long history --- studies such as \cite{jokisch2000learning} and \cite{chen1998rnn} employed objective metrics to assess prosodic features, such as Root Mean Squared Error (RMSE) related to pitch, duration, and intensity.  Recent studies on prosody related to speech synthesis aim to incorporate more objective metrics to perform a comprehensive acoustic analysis to evaluate quality, naturalness, and intelligibility. These metrics complement the predominant — and costly — subjective evaluation method used in TTS: the Mean Opinion Score (MOS) \cite{chan24_speechprosody}.

The appropriate use of pauses and pause duration, naturally employed by human speakers, enhances speech intelligibility by aiding in the interpretation of speech meaning \cite{Liu_2022}. In addition, pauses mark the boundaries of intonation groups and can align with syntactic boundaries within and between utterances \cite{viola08_speechprosody}. Along with pauses, the lengthening of the speech rate at the end of a segment, combined with acceleration at the beginning, known as discontinuities in the speech rate, serves as cues to identify boundaries and facilitates natural turn-taking in spontaneous conversations \cite{Biron_etal_2021}. These boundaries can be delineated by intonation units (IUs), which are segmentation units defined based on prosodic features \cite{Santos_etal_2022}. Among the various theoretical perspectives for analyzing these units, \cite{raso2012corpus} divides them into non-terminal breaks (NTB) and terminal breaks (TB), where Terminal Boundaries mark the conclusion of the utterance and Non-Terminal Boundaries mark breaks of non-conclusive sequences of the utterance.

This study examines the influence of manual and automatic annotations of prosodic boundaries in Brazilian Portuguese on the quality of synthesized speech.  In particular, it assesses the capacity of a non-autoregressive (NAR) TTS  model, FastSpeech 2\cite{ren2022fastspeech2fasthighquality}, to capture prosodic information when trained on a dataset manually segmented according to terminal intonation units. For comparison purposes, the same model is trained on segments automatically generated and transcribed by WhisperX, without subsequent human revision. This work seeks to address the current gap in the evaluation of prosodic annotations for Brazilian Portuguese and to provide results that may support future research in the training of spontaneous style text-to-speech synthesis models.

The key contributions of this work are as follows: (1) we present a comprehensive comparison between manual and automatic segmentation techniques, highlighting certain limitations of automatic segmentation; (2) we demonstrate the benefits and importance of prosodic segmentation for spontaneous speech synthesis; and (3) our code, datasets, and checkpoints are publicly available\footnote{It will be made publicly available once the blind-review process is complete}.

\section{NURC-SP Minimal Corpus}
To evaluate the impact of prosodic segmentation on spontaneous speech synthesis, we selected the Brazilian Portuguese NURC-SP Minimal Corpus (NURC-SP MC) dataset \cite{Santos_etal_2022}, which was manually annotated by human evaluators. Additionally, we created a version of the dataset that was automatically segmented and transcribed (Section \ref{sec:dataprocessing}).

NURC-SP MC \cite{Santos_etal_2022} is composed of 21 pairs of audio-transcription, although one of them (SP\_D2\_062) was removed from the dataset used in this study as its audio quality was not good (the voice is shaky) for training TTS.  The remaining pairs of NURC-SP MC are composed of: (i) 6 Formal Elocutions (EF); (ii) 5 formal dialogues between the speakers, with the presence of a documenter (D2); (iii) 9 interviews about different topics, carried out by an interviewer with the interviewee (DID).  NURC-SP MC was chosen because of its size, variety of text genres, and quality of annotation. The team of 6 linguists who annotated the dataset underwent annotation training, carried out in Praat \cite{Boersma_Weenink_2023}, and the inter-rater reliability on prosodic segmentation reached a kappa value above 0.8, indicating substantial agreement in terms of consistent annotation.

Although speech segmentation can be performed at different units (phones, syllables, stress groups, IUs, utterances, turns, and even larger domains), the adoption of a segmentation unit that takes into account prosodic features, such as the IU, has been a common practice in spontaneous speech corpora of several languages \cite{Biron_etal_2021}.  
Although IUs can be analysed from different theoretical perspectives, the boundaries of IUs are associated, in most definitions, with a set of typical acoustic features: (i) a coherent and unified pitch contour; (ii) boundary tones; (iii) pitch reset or pitch register change; (iv) pause; (v) one or more final syllable lengthening; (vi) change in speech rate at the end or beginning of a unit; and (vii) intensity changes \cite{Santos_etal_2022}.

The concept of intonation unit used in NURC-SP MC follows the prosodic segmentation model established by C-ORAL-BRASIL\footnote{https://www.c-oral-brasil.org/}. Accordingly, NURC-SP MC is annotated with prosodic boundaries (or breaks) marked as either terminal or non-terminal as already defined in Section \ref{sec:1}. It is important to notice that non-terminal prosodic breaks signal a non-autonomous prosodic unit inside the same utterance.

First, automatic forced alignment between the audio and the original transcription was performed using aeneas\footnote{https://github.com/readbeyond/aeneas},
which requires segmenting the text into excerpts. Thus, initially the text was segmented using the original annotation for pause, indicated by ellipsis in the NURC project\footnote{https://nurc.fflch.usp.br/}, and turn-shifts between speakers, indicated by a line break followed by the next speaker’s identification abbreviation. Then, the identification of prosodic breaks was based on the perceptual (auditory) relevance of prosodic cues, but the annotators were also oriented to rely on the visual inspection\footnote{The oscillogram and the spectrogram with the blue line showing the variation of the F0 curve throughout the recording were used by the annotators.} of the acoustic signal provided by Praat. The main cues to a prosodic break in Brazilian Portuguese are pause insertion, changes related to fundamental frequency (F0), and duration. 

It is important to notice that there are some minor differences from the definitions used in CORAL-BRASIL I, as NURC-SP MC did not make a distinction between interrupted, abandoned utterances (called a false start) and regular, concluded TBs. Furthermore, unlike CORAL-BRASIL I, NURC-SP MC segmented laughs as separate units. On the other hand, filled pauses (such as ``uh'', ``ah'', ``hmm''), false starts phenomena, and truncated words, for example, were not segmented into separate units as is the case with C-ORAL-BRASIL I.

\begin{figure*}[!htbp]
\centering
\includegraphics[width=\linewidth]{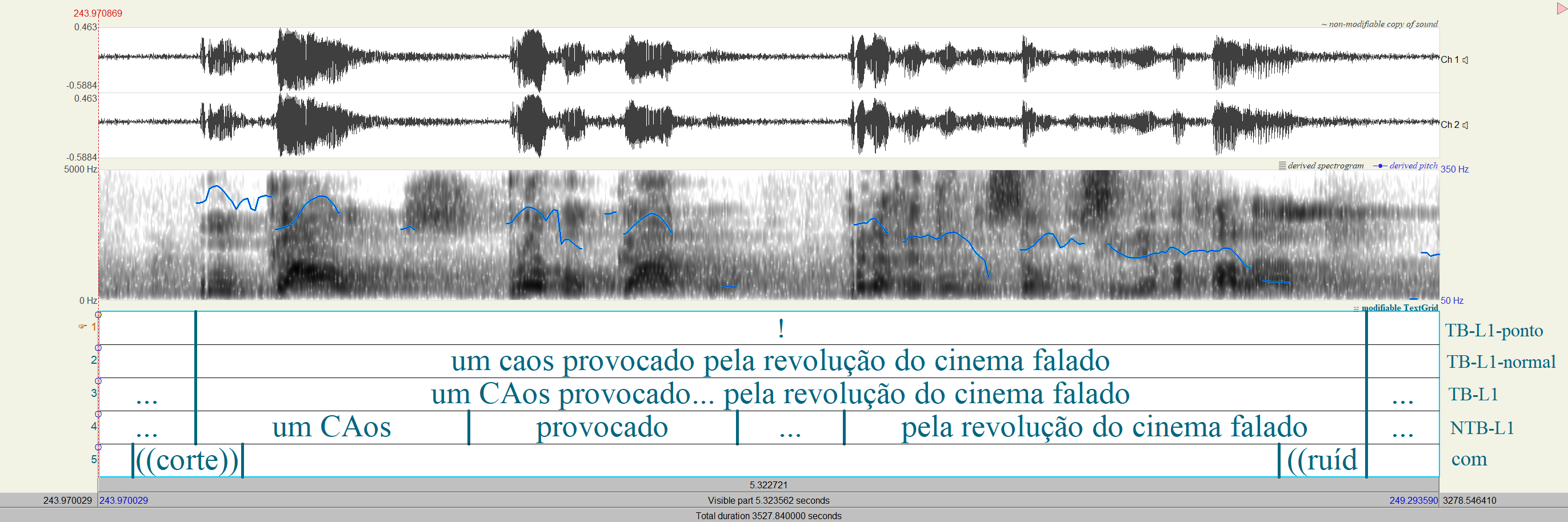}
\caption{Excerpt from the SP\_EF\_153 inquiry with five layers annotated in Praat: the first layer is used to indicate the punctuation that ends each TB (. ? ! … ), the second contains the normalized excerpt, i.e. without the annotation used for transcription in the NURC project, the next two for each speaker that appears in the inquiry (TB-L1, NTB-L1) and the last one for comments on the audio recording (com) \cite{Santos_etal_2022}.}
\label{fig:praat}
\end{figure*}

NURC-SP MC uses multilevel transcriptions that consist of the following interval layers annotated in the speech analysis program Praat. See Figure \ref{fig:praat} for an illustration of five layers in Praat:
(i) 2 layers for TB and NTB in which the speech of each main speaker (-L1, -L2) and documenter (-Doc1, -Doc2) is segmented into prosodic units and transcribed according to standards adapted from the NURC project; (ii) 1 layer (LA) for transcribed and segmented speech from any additional speaker; (iii)  1 layer for comments regarding the audio recording; (iv) 1 layer containing the normalized (-normal) version of the transcript of all TB and LA layers; and (v) 1 layer containing the punctuation (-point) that ends each TB (. ? ! …).

\section{Experiments}

Our experiments address the following questions:
(i)  whether there are statistically significant differences between natural and synthesized speech from models trained with the prosodic and automatic datasets, via acoustic features, including fundamental frequency (F0); 
(ii)  whether there are differences between manual and automatic segmentation (the average number of segments, the average size of segments, among others); and (iii)  whether there are differences in intelligibility of speech synthesized by models trained with the prosodic and automatic datasets, via Word Error Rate (WER) and Character Error Rate (CER).

We employed the manually annotated NURC-SP MC dataset and created an automatic version of the dataset using the WhisperX ASR model. WhisperX is a system designed for efficient transcription of long-form audio, providing accurate word-level timestamps. It was selected due to its state-of-the-art performance in automatic speech recognition and speech segmentation. Moreover, WhisperX has recently been used in the development of several open-source datasets \cite{evaldo-leal-etal-2025-mupe,he2024emilia,liu2025voxpopulitts}.

\subsection{Datasets Preprocessing}\label{sec:dataprocessing}

We duplicated the 20 NURC-SP MC audio files to segment and transcribe them using WhisperX  as well, and named this dataset ``automatic''\footnote{It is available at (omitted due to blind review)}. 

The original NURC-SP MC, which was manually transcribed in the 1970s, was revised before the manual prosodic annotation of segmentation was undertaken; in this study, it was called ``prosodic''\footnote{It is available at (omitted due to blind review)}.

We split both subsets (automatic and prosodic) into three sets: 

\begin{enumerate}
\item 7527 segments from 19 inquiries (about 11 hours and 50 minutes) for training the \textbf{prosodic model} and  9870 segments from 19 inquiries (about 15 hours and 50 minutes) for training the \textbf{automatic model};
\item 473 segments from the same 19 inquiries (about 42 minutes) for validation of \textbf{prosodic model} and 500 segments from the same 19 inquiries (about 44 minutes) for validation of the \textbf{automatic model}; and
\item  one inquiry (DID\_234) was separated for testing both models. Specifically, we selected 30 segments of the speaker L1 of DID\_234 for the acoustic analysis (Section \ref{sec:52}) and all the segments of the speaker L1 of DID\_234 (361 segments) for evaluating intelligibility (Section \ref{sec:51}). 
\end{enumerate}

The selection of segments for the validation set respected the duration intervals (short, medium, and long audios) and the representativeness of the inquiries.

To compose the subset called ``prosodic'', it was used only terminal intonation units (terminal boundaries mark the conclusion of the utterance) from NURC-SP MC. 
For the subset called ``automatic'', the 20 audio files of NURC-SP MC were segmented and transcribed by WhisperX. 

The prosodic subset has \textbf{12:32:25 hours} , and the automatic subset has \textbf{16:33:54 hours}. The difference (4 hours, 1 minute, 29 seconds) is due to three factors: (i) the header\footnote{The header includes information about inquiries (e.g. audio duration, recording date) and about main speakers (e.g. gender, age).} of each file that was preserved in the automatic subset; (ii) transcriptions containing only unintelligible segments and emotion sounds (e.g., laughter) were excluded from the prosodic subset; (iii) segments with speaker overlap of more than 2 seconds were removed from the prosodic subset; as well when the overlapping was greater than 50\% of the total time. The quality of intelligibility was also verified via transcription using Whisper, and segments with a CER > 0.6 were removed.
In our preliminary experiments, we observed that neither the ``automatic'' nor the ``prosodic'' subsets alone were sufficient to achieve convergence in the TTS models. We believe this is due to the size and characteristics of the dataset. To enable successful training, we included the Portuguese subset of the CML-TTS \cite{oliveira_2023_cmltts} dataset in the training data for both the ``automatic'' and ``prosodic'' subsets. CML-TTS  comprises audiobooks in seven languages: Dutch, French, German, Italian, Portuguese, Polish, and Spanish. For this work, only the Portuguese audio files were utilized, resulting in 59 hours (30 speakers) of data after the pre-processing.

\subsection{TTS Experiments}

In this study, we adopted the FastSpeech 2 TTS model \cite{ren2022fastspeech2fasthighquality}. FastSpeech 2 is a transformer-based non-autoregressive model with a simple structure that generates greater variation in speech by explicitly modeling three prosodic features: duration, pitch, and energy. Specifically, duration, pitch, and energy are extracted from the speech waveform and directly used as conditional inputs during training; during inference, the model utilizes the predicted values. We selected this model because it explicitly models prosodic features through dedicated pitch, energy, and duration predictors, enabling the manipulation of these features during inference. We employed an open-source implementation of the model\footnote{https://github.com/ming024/FastSpeech2}, and the training procedure follows a pipeline consisting of the following steps:

\begin{enumerate}
\item Prepare align: Audio files and transcriptions for each segment were extracted from original datasets and saved in .lab and .wav files inside a folder named after each speaker code;
\item MFA Align 1: The tool Montreal Forced Aligner\footnote{https://montreal-forced-aligner.readthedocs.io/en/latest/} was applied to transform graphemes to phonemes using the pretrained model portuguese\_brazil\_mfa and a lexicon dictionary was generated in a text file;
\item MFA Align 2: MFA align command was used to process the raw data and generate the alignments in TextGrid format;
\item From that, a pre-processing script generated estimated values for energy and pitch, a json file with speakers codes and a train/validation text files with phonetic transcriptions. Validation subset was a selection of 1\% random segments;
\item Finally, the model was trained with the preprocessed data on a RTX 4070 GPU for 720k steps (in approximately 4 days).
\end{enumerate}

For a fair comparison, this pipeline was executed twice: once for the CML-TTS subset combined with the NURC-SP MC ``prosodic'' dataset, and once for the CML-TTS subset combined with the NURC-SP MC ``automatic'' dataset. Since the implemented pipeline already included an automatic train-dev split, we used all segments from our dataset as input. \textbf{Audio samples}\footnote{\url{https://sites.google.com/view/bracis25-prosod-blind-review/home}} from both experiments are available for evaluation.

\section{Evaluation of Prosodic and Automatic Segmentation Methods}
We compared the NURC-SP MC ``prosodic'' and NURC-SP MC ``automatic'' datasets to identify the weaknesses of automatic segmentation. First, we pre-processed both datasets as described in Section~\ref{sec:prep}. Subsequently, we conducted several statistical analyses to highlight the differences between the two segmentation methods, as presented in Section~\ref{sec:stat}.

\subsection{Preprocessing and Alignment}
\label{sec:prep}

To simplify the analysis of segmentations, both human and automatic transcriptions underwent a normalization process. Initially, the text was converted to lowercase, and punctuation was removed. Subsequently, numbers were converted to their full textual form using the python library num2words, and whitespaces were removed. Following this, all segments were written into two text files, word by word, with the special token [seg] inserted at the location of each segmentation (manual or automatic). As WhisperX’s audios contain additional headers for metadata extraction, such headers were removed from automatic transcriptions for a fair comparison with manual transcriptions. 

For segment alignment, we used Python's ndiff function\footnote{https://docs.python.org/3/library/difflib.html}. This is equivalent to organizing transcription tokens, including segmentation labels, line by line, and executing the diff command used in version control management systems for source code, such as GIT\footnote{https://git-scm.com/}.  This allowed us to align both segmentation lists, capturing the number of differences, insertions, and deletions.  Based on this output, we were able to identify the matches and mismatches between both automatic and prosodic segmentation boundaries and to be able to calculate the metrics mentioned previously. This approach enabled a detailed analysis of the alignment between segments and quantitative insights into each boundary strategy.

\subsection{Statistics of Segmentation}
\label{sec:stat}

Table \ref{tab:TokenandSegmentStatistics} presents statistics regarding the segmentation process. In total, the automatic process resulted in 130,220 tokens while the manual (prosodic) segmentation totaled 112,023 tokens. This difference is mainly due to the removal of 1,750 segments from the prosodic version, due to laughter, voice overlap or unintelligible segments (see Section \ref{sec:dataprocessing}), although other factors also play a role in this difference, for example, different styles of transcription for disfluencies (e.g.: word repetitions and hesitations) and minor transcription errors in the WhisperX's transcriptions\footnote{Headers present in WhisperX's transcriptions were removed in this version, as indicated in Section \ref{sec:prep}}.

\begin{table}[htb]
\centering
\caption{Token and Segment Statistics}
\label{tab:TokenandSegmentStatistics}
\begin{tabular}{lrrr}
\hline
\multicolumn{1}{l}{\textbf{Metrics}} & \textbf{Auto} & \textbf{Prosodic} & \textbf{Diff} \\ \hline
Total Tokens & 130,220 & 112,023  & 18,197 \\ \hline
Total Segments & 10,182 & 7,816  & 2,366 \\ \hline
Average Tokens per Segment & 12.79 ($\pm$ 9.69) & 14.33 ($\pm$ 13) & 2.14  \\ \hline
Average Segments per Interview  & 536 ($\pm$  233.18) & 411 ($\pm$  232.78) & 125 \\ \hline
\end{tabular}
\end{table}

Our results also show a difference in the number of total segments. Manual transcriptions use fewer segments ($\approx 7k$) than WhisperX's ones ($\approx 10k$), even considering the removal of 1,750 from the prosodic version. 
It is important to note that manual transcriptions in this work use only TB's. This number should be higher if we were using NTB instead (see Figure \ref{fig:praat}).
Also, annotators do not face WhisperX's restriction on audio length uniformity (maximum 30 seconds), which may also result in a slight reduction in the total number of segments.
More importantly, the automatic method generates 12.79 tokens per segment, on average, while the prosodic method generates 14.33 segments. This indicates that the prosodic segments tend to be larger. 
This is partially explained by disfluencies in manual transcriptions and errors in automatic transcriptions. However, the main factor is that manual transcriptions are longer, as TBs focus on complete utterances, capturing more information, without being restricted by maximum length constraints.

\begin{figure}
    \centering
    \includegraphics[width=0.8\linewidth]{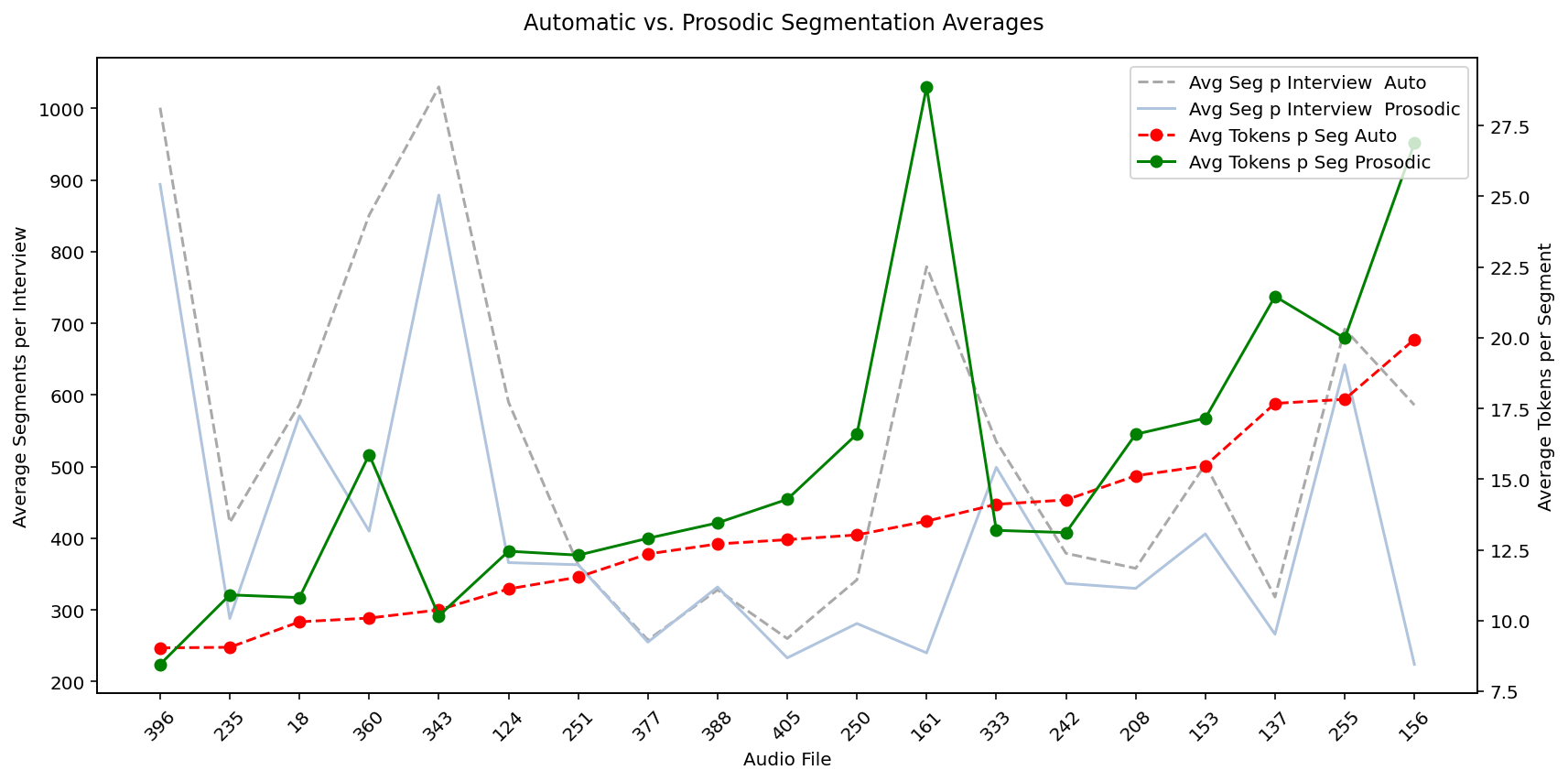}
    \caption{Automatic vs. Prosodic: Averages by Token and Segment}
    \label{fig:average-indicatorII}
\end{figure}

Figure \ref{fig:average-indicatorII} presents the average tokens per segment of the automatic and prosodic segmentations. 
Although WhisperX's segmentation was not originally conceived or trained to detect prosodic boundaries, it is useful to compare how well WhisperX's segments align with prosodic segments. To do so, we considered prosodic segment boundaries as ground truth labels, and WhisperX's segments as boundary predictions. We considered that a segment from WhisperX is aligned with a segment prosodic boundary when they both occur between a given pair of words, rather than looking for a precise timestamp in the audio.
Therefore, this problem can be thought of as a binary classification problem that indicates whether there is a segment boundary between any pair of words, which allowed us to calculate precision, recall, and F1-score for the positive class (the presence of a boundary). 

\begin{figure}[htb]
    \centering
    \includegraphics[width=1.0\linewidth]{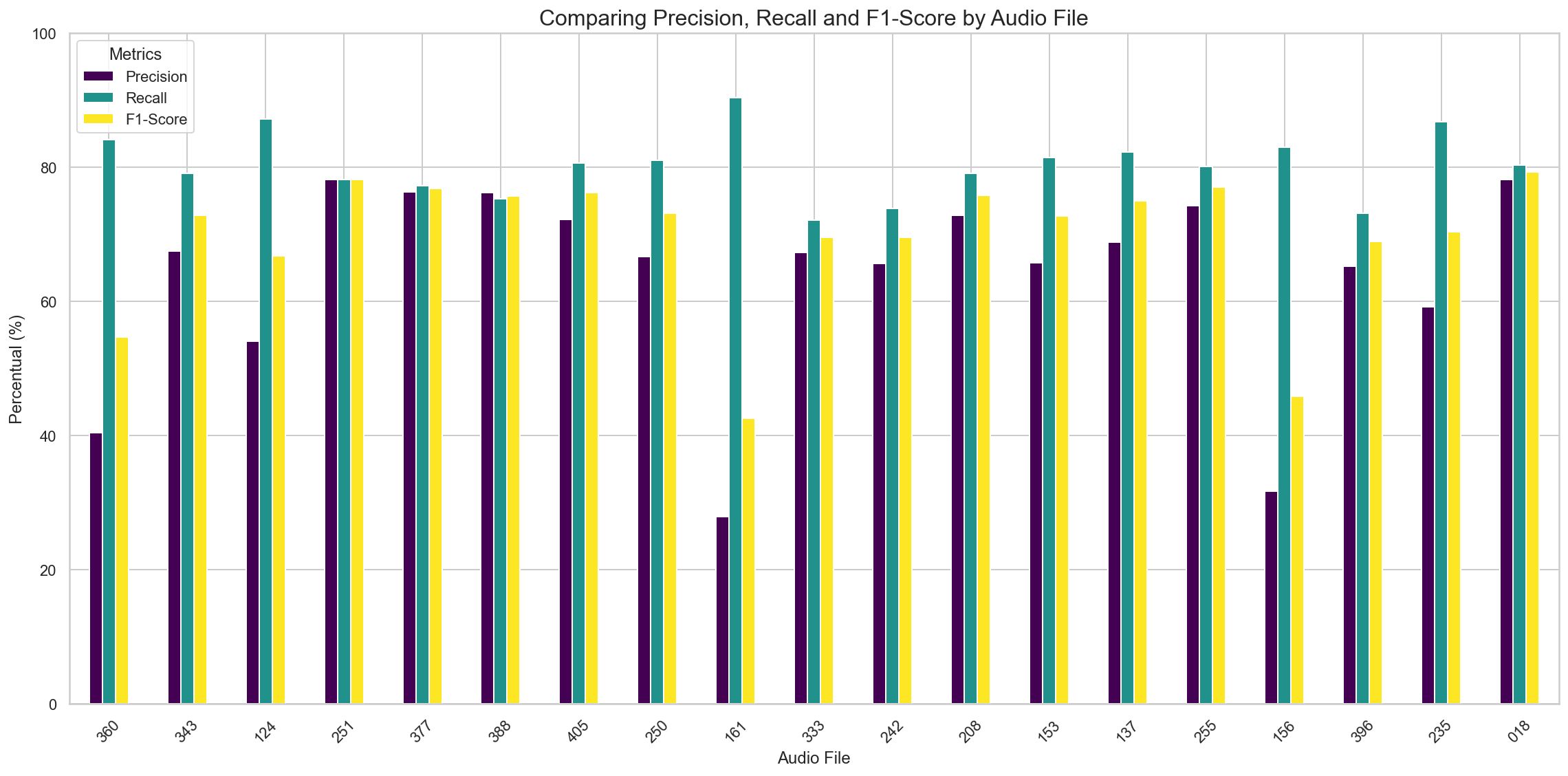}
    \caption{Precision, Recall, and F1-Score by Inquiry.}
    \label{fig:comparingmetricsprecisionrecallf1score}
\end{figure}

As a result, we obtained a precision of 60.95\% and a recall of 79.40\%, resulting in an F1-score of 68.96\%. In this context, precision increases when automatic segments do not occur outside prosodic boundaries. Similarly, recall increases when automatic segments do occur within prosodic boundaries. The fact that recall is higher than precision indicates that WhisperX tends to segment at prosodic boundaries. The smaller precision also indicates that WhisperX inserts more boundaries than the manual process, for the reasons previously discussed. Figure \ref{fig:comparingmetricsprecisionrecallf1score} presents individual metrics for each inquiry. It should be noticed that there is a small variation in automatic segmentation (red line), while prosodic segmentation varies more. This result is expected because prosodic segmentation focuses on information units rather than on audio length.

\section{Results of TTS Intelligibility and Acoustic Analysis of Prosodic Features}
\label{sec:5}

\subsection{Intelligibility}
\label{sec:51}
To evaluate the intelligibility of the synthesized audios, we used an ASR model\footnote{https://huggingface.co/nilc-nlp/distil-whisper-coraa-mupe-asr} that uses two large spontaneous speech datasets in Brazilian Portuguese to transcribe them and calculate CER and WER with the human-annotated labels. For comparison, we also transcribed the original test audios with the same model. These metrics are commonly used in the TTS field \cite{casanova2024xtts,hussain2025koel,kim2022guided} and provide insights into pronunciation accuracy, reflecting the impact of the segmentation process.

A total of 361 segments were synthesized for the speaker L1 of DID\_234 (SP-DID-234-TB-L1) with the automatic and prosodic FastSpeech 2 checkpoints, ending up with 1083 total audios (including the segments of the ground truth original). The CER/WER averages for the three audios for each segment 
ground CER = 0.09, ground WER = 0.16, auto CER = 0.35, auto WER = 0.50, prosodic CER = 0.31, prosodic WER = 0.43.
The prosodic values were slightly better than automatic ones (we applied T-Test for auto and prosodic averages, ending with WER(t=2.589, p-value<0.01) and CER(t=1.796, p-value=0.07). Although the CER average was not statistically significant, it is close to the 0.05 threshold. 

The CSV file with all transcriptions and individual CER/WER is available at our GitHub repository (omitted due to blind review).

\subsection{Acoustic Analysis of Prosodic Features}
\label{sec:52}
Regarding the acoustic analysis, we used a sample of natural speech for the testing phase of our experiment in order to prosodically compare it with the utterances synthesized by FastSpeech 2. We selected the speaker SP-DID-234-TB-L1, as it represents a DID (Dialog between Informant and Documenter) recording, providing high-quality material for prosodic analysis and utterance selection. This speaker also presented the largest number of segments among all samples belonging to the DID genre. For the purposes of this study, we selected 30 neutral declarative utterances, free from emphasis, turn-taking overlap, laughter, noise, or other paralinguistic elements. Each utterance contained either a single terminal break (TB) or a non-terminal break (NTB) within a TB. Selecting neutral declarative utterances from a spontaneous speech corpus is a challenging task; therefore, 30 utterances were the maximum number closely matching the established criteria

For the prosodic annotation, four points on the F0 contour were annotated in each of the 30 utterances, based on both existing literature and visual inspection of the contour. Specifically, the onset point (ons) corresponds to the first syllable of the intonational unit; the pre-nuclear point (prnuc) marks the lowest F0 value before the nuclear accent; the nuclear point (nuc) identifies the highest F0 peak, typically associated with neutral declarative utterances in Brazilian Portuguese; and the post-nuclear point (psnuc) corresponds to the lowest movement during the final F0 fall. These points were selected to examine the F0 curve in natural speech and to compare it with the corresponding curve in synthesized speech.

\begin{figure}[htb]
\caption{Average of four F0 contour points in the original audio and trained TTS models}
\label{fig:fourpoints}
\centering
\includegraphics[scale=0.5]{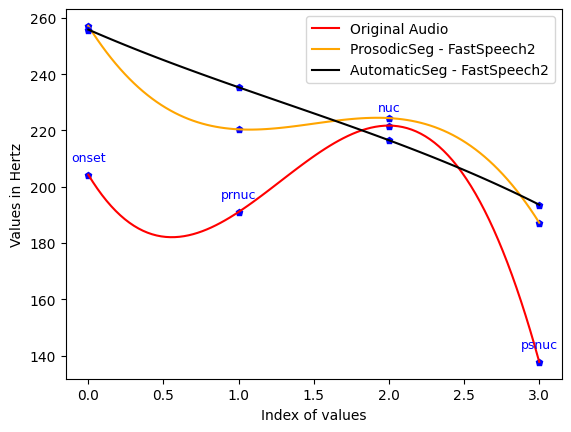}
\end{figure}

For this purpose, we extracted the maximum F0 value at each annotated point using the AnalyseTier script \cite{hirst2012analyse}, following standard practices in the literature \cite{swerts1997prosodic}. The extracted values were exported to a spreadsheet, and the mean F0 values for each point, in both natural and synthesized speech, were calculated to support the subsequent statistical analysis (see Figure \ref{fig:fourpoints}).

We calculated how closely each FastSpeech 2 training approach approximated natural speech, using root mean square error (RMSE) to measure the discrepancy between the observed values in the curve. The higher the numerical value of the model, the farther it is from the original audio. The RMSE calculation resulted in a value of approximately 39.07 Hz between FastSpeech 2 with prosodic segmentation and natural speech, which means that the curve generated by FastSpeech 2, after being trained with a manually segmented dataset, is relatively closer to the natural one when compared to the curve generated by FastSpeech 2 with automatic segmentation, which is slightly more distant from original speech (RMSE = approximately 44.05 Hz). To compare the average values of different prosodic points among the three groups, we performed separate univariate ANOVAs. The results showed a significant difference for onset (p-value<0.01), pre-nuclear (p-value=0.05), and post-nuclear (p-value<0.01) points, but no significant difference for the nuclear point (p-value=0.86). We also conducted a Tukey post-hoc test to explore which groups differed from each other and at which points. For the onset variable, both FastSpeech 2 training versions significantly differed from natural speech, while the two model versions were closer to each other. This means that the difference between the model and natural speech is noticeable right at the beginning of the utterances. 

In the pre-nuclear position, there was no significant difference between the original audio and FastSpeech 2 with prosodic segmentation (p-value=0.08), making it closer to natural speech at this point than FastSpeech 2 with automatic segmentation (p-value=0.04). At the nuclear point, there was no significant difference between the three groups, as previously indicated by the ANOVA — in other words, no significant difference between natural speech and the two FastSpeech 2 training versions. This shows that the model was able to reproduce the nuclear accent that characterizes declarative utterances in Brazilian Portuguese \cite{moraes08_speechprosody}. The post-nuclear point differed significantly between both FastSpeech 2 models and natural speech. Even though there was a statistical difference between the model trainings and natural audio, it is important to consider the overall behavior of the intonational curve. From a linguistic perspective, the shape of the curve may matter more than just the individual average values of the four points. A visual interpretation shows that FastSpeech 2 with prosodic segmentation produces a curve with a low pre-nuclear point that precedes the rise of the nuclear accent peak, just like natural speech, while FastSpeech 2 with automatic segmentation shows a simple downward linear slope.

We also examined the average F0 variation of the utterances in semitones (relative to 100 Hz), 
based on the maximum F0 values. The ANOVA result indicated that there was a significant difference between the groups (p-value<0.01), then we perform a post-hoc test. While the two FastSpeech 2 training methods did not differ from each other (p-value=0.99), neither was able to approximate natural speech (Prosodic segmentation = p-value<0.01, Automatic segmentation = p-value<0.01). Humans discriminate F0 variation more effectively when it is measured in semitones, since we are more sensitive to changes in lower frequencies than in higher ones \cite{barbosa2019prosodia}. These results regarding F0 variation can objectively explain why this speech sounds less natural.

\section{Conclusions}

This study evaluated the impact of manual versus automatic prosodic boundary segmentation in Brazilian Portuguese on the quality of synthesized spontaneous speech. The main objective was to determine whether incorporating human-annotated prosodic boundaries into a TTS training corpus could improve the performance of an NAR TTS model compared to using automatically segmented data. By comparing these two approaches, we aimed to assess differences in intelligibility and how closely the synthetic speech aligns with natural prosodic patterns. 

The results indicate that models trained with manually segmented prosodic data achieved slightly better performance than those trained with automatic segmentation. In particular, the FastSpeech 2 model trained on the prosodic dataset produced speech that was marginally more intelligible (as evidenced by a lower WER/CER) and acoustically closer to natural speech. For example, acoustic analyses of intonation contours showed that the prosody-informed model more accurately reproduced natural pitch patterns in neutral declarative utterances. In contrast, the model trained on automatically segmented data (using WhisperX) yielded more uniform speech segments and a flatter overall intonation, missing some of the expressive variability present in natural prosody. These findings suggest that explicit prosodic boundary annotation provides a modest but measurable advantage in capturing the nuances of spontaneous speech. 
Notably, the automatic segmentation method demonstrated promising performance in detecting prosodic breaks. The WhisperX-based pipeline achieved high recall of the boundaries marked by human annotators, illustrating its potential for efficient corpus building. However, there remains a noticeable gap between the two segmentation approaches in terms of speech naturalness. The automatically segmented dataset did not fully capture the variability and subtle prosodic cues that human annotation provided. As a result, the synthetic speech from the automatic approach sounded less dynamic and natural compared to the speech from the prosodically segmented dataset. This gap highlights that current automatic methods, despite their convenience, cannot yet replicate the rich prosodic patterns that come from careful manual segmentation. This study reinforces that careful evaluation of prosodic and acoustic characteristics is crucial for advancing automatic speech synthesis systems. Analyses of features such as pitch contours
provided objective evidence of how segmentation strategies influence the naturalness and intelligibility of the output. Incorporating such evaluations into the development cycle of TTS systems is essential to bridge the gap between synthetic and natural speech, particularly in applications that demand conversational fluency and expressive variability.

This work has certain limitations that must be acknowledged. The scope of our experiments was restricted to a relatively small dataset of Brazilian Portuguese spontaneous speech, which may limit the generalizability of the conclusions to other languages or speaking styles. Additionally, the automatic segmentation using WhisperX, while effective for generating aligned transcripts, is not specifically tailored to prosodic boundary detection; this limitation likely affected the quality of the automatically segmented training data. Future efforts will focus on expanding and diversifying prosodically annotated corpora, which would provide more data for training and enable a deeper analysis of prosody in TTS. We also plan to refine automatic segmentation techniques — for instance, by incorporating acoustic-prosodic cues or developing dedicated prosody segmentation models — to narrow the gap between automatic and manual segmentation performance. Indeed, there are some methods that explicitly model speech prosody to build automatic prosody annotation for Chinese \cite{dai22_interspeech} and English \cite{zhong24c_interspeech}.
Moreover, it will be important to evaluate the benefits of prosodic segmentation in different conditions, including other genres of speech, to verify that the advantages observed here extend beyond Brazilian Portuguese. Additionally, we plan to expand our study to include other TTS models, such as autoregressive and flow-matching-based architectures.

\bibliographystyle{splncs04}
\bibliography{paper}

@inproceedings{dai22_interspeech,
  title     = {Automatic Prosody Annotation with Pre-Trained Text-Speech Model},
  author    = {Ziqian Dai and Jianwei Yu and Yan Wang and Nuo Chen and Yanyao Bian and GuangZhi Li and Deng Cai and Dong Yu},
  year      = {2022},
  booktitle = {Interspeech 2022},
  pages     = {5513--5517},
  doi       = {10.21437/Interspeech.2022-10005},
  issn      = {2958-1796}
}

@inproceedings{chan24_speechprosody,
  title     = {Exploring the accuracy of prosodic encodings in state-of-the-art text-to-speech models},
  author    = {Cedric Chan and Jianjing Kuang},
  year      = {2024},
  booktitle = {Speech Prosody 2024},
  pages     = {27--31},
  doi       = {10.21437/SpeechProsody.2024-6},
  issn      = {2333-2042},
}

@inproceedings{bain2022whisperx,
  title     = {WhisperX: Time-Accurate Speech Transcription of Long-Form Audio},
  author    = {Max Bain and Jaesung Huh and Tengda Han and Andrew Zisserman},
  year      = {2023},
  booktitle = {Interspeech 2023},
  pages     = {4489--4493},
  doi       = {10.21437/Interspeech.2023-78},
  issn      = {2958-1796},
}

@inproceedings{liu2025voxpopulitts,
  title={VoxpopuliTTS: a large-scale multilingual TTS corpus for zero-shot speech generation},
  author={Liu, Wenrui and Bai, Jionghao and Cheng, Xize and Zuo, Jialong and Jiang, Ziyue and Ji, Shengpeng and Fang, Minghui and Yang, Xiaoda and Yang, Qian and Zhao, Zhou},
  booktitle={Proceedings of the 31st International Conference on Computational Linguistics},
  pages={10293--10297},
  year={2025}
}

@inproceedings{kim2022guided,
  title={Guided-tts: A diffusion model for text-to-speech via classifier guidance},
  author={Kim, Heeseung and Kim, Sungwon and Yoon, Sungroh},
  booktitle={International Conference on Machine Learning},
  pages={11119--11133},
  year={2022},
  organization={PMLR}
}

@article{hussain2025koel,
  title={Koel-TTS: Enhancing LLM based Speech Generation with Preference Alignment and Classifier Free Guidance},
  author={Hussain, Shehzeen and Neekhara, Paarth and Yang, Xuesong and Casanova, Edresson and Ghosh, Subhankar and Desta, Mikyas T and Fejgin, Roy and Valle, Rafael and Li, Jason},
  journal={arXiv:2502.05236},
  year={2025}
}

@inproceedings{casanova2024xtts,
  title={XTTS: a Massively Multilingual Zero-Shot Text-to-Speech Model},
  author={Casanova, Edresson and Davis, Kelly and G{\"o}lge, Eren and G{\"o}knar, G{\"o}rkem and Gulea, Iulian and Hart, Logan and Aljafari, Aya and Meyer, Joshua and Morais, Reuben and Olayemi, Samuel and others},
  booktitle={Proc. Interspeech 2024},
  pages={4978--4982},
  year={2024}
}

@inproceedings{Santos_etal_2022,
  title        = {{CORAA NURC-SP Minimal Corpus}: a manually annotated corpus of {B}razilian {P}ortuguese spontaneous speech},
  author       = {Santos, Vinícius G. and Caroline Adriane Alves and Bruno Baldissera Carlotto and Bruno Angelo Papa Dias and Lucas Rafael Stefanel Gris and Renan de Lima Izaias and Maria Luiza Azevedo de Morais and Paula Marin de Oliveira and Rafael Sicoli and Flaviane Romani Fernandes-Svartman and Marli Quadros Leite and Sandra Maria Aluísio},
  year         = 2022,
  booktitle    = {Proc.\ {IberSPEECH} 2022},
  pages        = {161--165},
  doi          = {10.21437/IberSPEECH.2022-33}
}

@inproceedings{he2024emilia,
  title={Emilia: An extensive, multilingual, and diverse speech dataset for large-scale speech generation},
  author={He, Haorui and Shang, Zengqiang and Wang, Chaoren and Li, Xuyuan and Gu, Yicheng and Hua, Hua and Liu, Liwei and Yang, Chen and Li, Jiaqi and Shi, Peiyang and others},
  booktitle={2024 IEEE Spoken Language Technology Workshop (SLT)},
  pages={885--890},
  year={2024},
  organization={IEEE}
}

@ARTICLE{Liu_2022,
AUTHOR={Liu, Shimeng and Nakajima, Yoshitaka and Chen, Lihan and Arndt, Sophia and Kakizoe, Maki and Elliott, Mark A. and Remijn, Gerard B.},    
TITLE={How Pause Duration Influences Impressions of English Speech: Comparison Between Native and Non-native Speakers},      
JOURNAL={Frontiers in Psychology},     	
VOLUME={13},           	
YEAR={2022},      	  
URL={https://www.frontiersin.org/articles/10.3389/fpsyg.2022.778018},       	
DOI={10.3389/fpsyg.2022.778018},      	
ISSN={1664-1078}
}

@book{sagisaka1997computing,
  title={Computing prosody: computational models for processing spontaneous speech},
  author={Sagisaka, Yoshinori and Campbell, Nick and Higuchi, Norio},
  year={1997},
  publisher={Springer Science \& Business Media}
}

@inproceedings{jokisch2000learning,
  title={Learning the parameters of quantitative prosody models},
  author={Jokisch, Oliver and Mixdorff, Hansj{\"o}rg and Kruschke, Hans and Kordon, Ulrich},
  booktitle = {6th International Conference on Spoken Language Processing (ICSLP 2000)},
  pages={645--648},
  year={2000},
  doi= {10.21437/ICSLP.2000-160},
  issn      = {2958-1796}
}

@article{chen1998rnn,
  title={An {RNN}-based prosodic information synthesizer for Mandarin text-to-speech},
  author={Chen, Sin-Horng and Hwang, Shaw-Hwa and Wang, Yih-Ru},
  journal={IEEE transactions on speech and audio processing},
  volume={6},
  number={3},
  pages={226--239},
  year={1998},
  publisher={IEEE},
  doi={10.1109/89.668817}
}

@inproceedings{viola08_speechprosody,
  title     = {The roles of pause in speech expression},
  author    = {Izabel Cristina Viola and Sandra Madureira},
  year      = {2008},
  booktitle = {Speech Prosody 2008},
  pages     = {721--724},
  doi       = {10.21437/SpeechProsody.2008-160},
  issn      = {2333-2042},
}

@book{raso2012corpus,
  title={O Corpus c-oral-brasil},
  author={Raso, Tommaso and Mello, Heliana},
  year={2012},
  publisher={Editora UFMG, Belo Horizonte},
}

@inproceedings{zhong24c_interspeech,
  title     = {Multi-Modal Automatic Prosody Annotation with Contrastive Pretraining of Speech-Silence and Word-Punctuation},
  author    = {Jinzuomu Zhong and Yang Li and Hui Huang and Korin Richmond and Jie Liu and Zhiba Su and Jing Guo and Benlai Tang and Fengjie Zhu},
  year      = {2024},
  booktitle = {Interspeech 2024},
  pages     = {2305--2309},
  doi       = {10.21437/Interspeech.2024-2133},
  issn      = {2958-1796},
}

@article{Biron_etal_2021,
  title        = {Automatic detection of prosodic boundaries in spontaneous speech},
  author       = {Tirza Biron and Daniel Baum and Dominik Freche and Nadav Matalon and Netanel Ehrmann and Eyal Weinreb and David Biron and Elisha Moses},
  year         = 2021,
  month        = 5,
  journal      = {PLoS ONE},
  publisher    = {Public Library of Science},
  volume       = 16,
  number       = 5,
  pages        = {1--21},
  doi          = {10.1371/journal.pone.0250969}
}

@misc{xie2025controllablespeechsynthesisera,
      title={Towards Controllable Speech Synthesis in the Era of Large Language Models: A Survey}, 
      author={Tianxin Xie and Yan Rong and Pengfei Zhang and Wenwu Wang and Li Liu},
      year={2025},
      eprint={2412.06602},
      archivePrefix={arXiv},
      primaryClass={cs.CL},
      url={https://arxiv.org/abs/2412.06602}, 
}

@inproceedings{evaldo-leal-etal-2025-mupe,
    title = "{M}u{P}e Life Stories Dataset: Spontaneous Speech in {B}razilian {P}ortuguese with a Case Study Evaluation on {ASR} Bias against Speakers Groups and Topic Modeling",
    author = "Leal, Sidney Evaldo and
      Candido Junior, Arnaldo  and
      Marcacini, Ricardo  and
      Casanova, Edresson  and
      Gon{\c{c}}alves, Odilon  and
      Silva Soares, Anderson  and
      Freitas Lima, Rodrigo  and
      Stefanel Gris, Lucas Rafael  and
      Alu{\'i}sio, Sandra",
    editor = "Rambow, Owen  and
      Wanner, Leo  and
      Apidianaki, Marianna  and
      Al-Khalifa, Hend  and
      Eugenio, Barbara Di  and
      Schockaert, Steven",
    booktitle = "Proceedings of the 31st International Conference on Computational Linguistics",
    month = jan,
    year = "2025",
    address = "Abu Dhabi, UAE",
    publisher = "Association for Computational Linguistics",
    url = "https://aclanthology.org/2025.coling-main.407/",
    pages = "6076--6087"
}

@inproceedings{radford23_whisper,
title = 	 {Robust Speech Recognition via Large-Scale Weak Supervision},
author =       {Radford, Alec and Kim, Jong Wook and Xu, Tao and Brockman, Greg and Mcleavey, Christine and Sutskever, Ilya},
booktitle = 	 {Proceedings of the 40th International Conference on Machine Learning},
pages = 	 {28492--28518},
year = 	 {2023},
editor = 	 {Krause, Andreas and Brunskill, Emma and Cho, Kyunghyun and Engelhardt, Barbara and Sabato, Sivan and Scarlett, Jonathan},
volume = 	 {202},
series = 	 {Proceedings of Machine Learning Research},
month = 	 {23--29 Jul},
publisher =    {PMLR}
}

@InProceedings{oliveira_2023_cmltts,
author="Oliveira, Frederico S.
and Casanova, Edresson
and Junior, Arnaldo Candido
and Soares, Anderson S.
and Galv{\~a}o Filho, Arlindo R.",
editor="Ek{\v{s}}tein, Kamil
and P{\'a}rtl, Franti{\v{s}}ek
and Konop{\'i}k, Miloslav",
title="CML-TTS: A Multilingual Dataset for Speech Synthesis in Low-Resource Languages",
booktitle="Text, Speech, and Dialogue",
year="2023",
publisher="Springer Nature Switzerland",
address="Cham",
pages="188--199",
isbn="978-3-031-40498-6"
}

@misc{Boersma_Weenink_2023,
  title        = {Praat: doing phonetics by computer [{C}omputer program]. {V}ersion 6.3.10},
  author       = {Boersma, Paul and Weenink, David},
  year         = 2023,
  url          = {http://www.praat.org/}
}

@article{swerts1997prosodic,
  title={Prosodic features at discourse boundaries of different strength},
  author={Swerts, Marc},
  journal={The Journal of the Acoustical Society of America},
  volume={101},
  number={1},
  pages={514--521},
  year={1997},
  publisher={Acoustical Society of America}
}

@misc{hirst2012analyse,
  title={Analyse tier PRAAT script},
  author={Hirst, Daniel},
  year={2012}
}

@misc{ren2022fastspeech2fasthighquality,
      title={FastSpeech 2: Fast and High-Quality End-to-End Text to Speech}, 
      author={Yi Ren and Chenxu Hu and Xu Tan and Tao Qin and Sheng Zhao and Zhou Zhao and Tie-Yan Liu},
      year={2022},
      eprint={2006.04558},
      archivePrefix={arXiv},
      primaryClass={eess.AS},
      url={https://arxiv.org/abs/2006.04558}, 
}

@inproceedings{li24na_interspeech,
  title     = {Spontaneous Style Text-to-Speech Synthesis with Controllable Spontaneous Behaviors Based on Language Models},
  author    = {Weiqin Li and Peiji Yang and Yicheng Zhong and Yixuan Zhou and Zhisheng Wang and Zhiyong Wu and Xixin Wu and Helen Meng},
  year      = {2024},
  booktitle = {Interspeech 2024},
  pages     = {1785--1789},
  doi       = {10.21437/Interspeech.2024-1989},
  issn      = {2958-1796},
}

@inproceedings{moraes08_speechprosody,
  title     = {The pitch accents in Brazilian portuguese: analysis by synthesis},
  author    = {João Antônio de Moraes},
  year      = {2008},
  booktitle = {Speech Prosody 2008},
  pages     = {389--397},
  doi       = {10.21437/SpeechProsody.2008-4},
  issn      = {2333-2042},
}

@book{barbosa2019prosodia,
  title={Pros{\'o}dia},
  author={Barbosa, Plinio Almeida},
  year={2019},
  publisher={Par{\'a}bola}
}

@inproceedings{matos-etal-2024-accent,
    title = "Accent Classification is Challenging but Pre-training Helps: a case study with novel {B}razilian {P}ortuguese datasets",
    author = "Matos, Ariadne  and
      Ara{\'u}jo, Gustavo  and
      Junior, Arnaldo Candido  and
      Ponti, Moacir",
    editor = "Gamallo, Pablo  and
      Claro, Daniela  and
      Teixeira, Ant{\'o}nio  and
      Real, Livy  and
      Garcia, Marcos  and
      Oliveira, Hugo Gon{\c{c}}alo  and
      Amaro, Raquel",
    booktitle = "Proceedings of the 16th International Conference on Computational Processing of Portuguese - Vol. 1",
    month = mar,
    year = "2024",
    address = "Santiago de Compostela, Galicia/Spain",
    publisher = "Association for Computational Lingustics",
    url = "https://aclanthology.org/2024.propor-1.37/",
    pages = "364--373"
}
\end{document}